\documentclass[11pt,a4paper]{article}
\usepackage[hyperref]{resources/acl2020}
\usepackage{times}
\usepackage{latexsym}
\usepackage{graphicx}
\graphicspath{ {./} }
\usepackage{url}
\usepackage{xspace}
\usepackage{comment}
\usepackage{multirow}
\usepackage{booktabs} 

\aclfinalcopy 

\newcommand{\tableRef}[1]{Table \ref{#1}}
\newcommand{\sectionRef}[1]{\S \ref{#1}}
\newcommand{\figureRef}[1]{Figure  \ref{#1}}

\newcommand{\percentage}[1]{$#1$\%\xspace}
\newcommand{\statistic}[1]{$#1$\xspace}

\newcommand{\supp}[0]{Appendix\xspace}

\newcommand{\paren}[1]{\left({#1}\right)}
\newcommand{\braces}[1]{\left\{{#1}\right\}}

\newcommand{\datasetURL}[0]{\footnote{\url{https://www.research.ibm.com/haifa/dept/vst/debating_data.shtml\#Debate Speech Analysis}}}

\newcommand{\vk}[1]{v_{{#1}|k}}

\newcommand{\pk}[0]{p\paren{k}}

\newcommand{\allSpeechesCriterion}[0]{All}
\newcommand{\explicitSpeechesCriterion}[0]{Explicit}
\newcommand{\implicitSpeechesCriterion}[0]{Implicit}

\newcommand{\annotationQuestionsAll}[0]{\statistic{60}}

\newcommand{\accuracyTitle}[0]{{\bf A}}
\newcommand{\randomTitle}[0]{\textbf{R}}
\newcommand{\precisionTitle}[0]{\textbf{A}}
\newcommand{\mrrTitle}[0]{\textbf{M}}

\newcommand{\supportingSpeechTitle}[0]{\textbf{\#Su}}
\newcommand{\numCandsTitle}[0]{\textbf{\#Op}}
\newcommand{\averageCandsTitle}[0]{\textbf{av. \#C}}
\newcommand{\percentPositiveTitle}[0]{\textbf{Pos}}

\newcommand{\crowdRowName}[0]{Cr\xspace}
\newcommand{\expertsRowName}[0]{Ex\xspace}

\newcommand{\figureEightName}[0]{Figure-Eight\xspace}

\newcommand{\implicitInternalPrecision}[0]{\statistic{76.4}}
\newcommand{\implicitInternalPrecisionPercentage}[0]{\percentage{76}}

\newcommand{\explicitInternalPrecision}[0]{\statistic{91.7}}
\newcommand{\explicitInternalPrecisionPercentage}[0]{\percentage{92}}

\newcommand{\allInternalPrecision}[0]{\statistic{85.6}}
\newcommand{\allInternalPrecisionPercentage}[0]{\percentage{86}}

\newcommand{\implicitInternalRandomPrecision}[0]{\statistic{30.2}}

\newcommand{\explicitInternalRandomPrecision}[0]{\statistic{37.4}}

\newcommand{\allInternalRandomPrecision}[0]{\statistic{31.4}}
\newcommand{\allInternalRandomPrecisionPercentage}[0]{\percentage{31}}

\newcommand{\explicitCrowdPrecision}[0]{\statistic{55.6}}

\newcommand{\implicitCrowdPrecision}[0]{\statistic{38.9}}

\newcommand{\allCrowdPrecision}[0]{\statistic{48.9}}

\newcommand{\explicitCrowdRandomPrecision}[0]{\statistic{29.5}}

\newcommand{\implicitCrowdRandomPrecision}[0]{\statistic{26.6}}

\newcommand{\allCrowdRandomPrecision}[0]{\statistic{28.4}}

\newcommand{\trainName}[0]{Train}
\newcommand{\validationName}[0]{Validation}
\newcommand{\testName}[0]{Test}

\newcommand{\trainNameLower}[0]{train\xspace}
\newcommand{\validationNameLower}[0]{validation\xspace}
\newcommand{\testNameLower}[0]{test\xspace}

\newcommand{\vanilaBert}{BERT\xspace}
\newcommand{\tunedBert}{BERT-T\xspace}
\newcommand{\simDissimName}{SD\xspace}
\newcommand{\simDissimEmbeddingsName}{SD-e\xspace}
\newcommand{\docSimCosine}{Cos\xspace}
\newcommand{\docSimJS}{JS\xspace}

\newcommand{\ngramScoreName}{ngrs\xspace}
\newcommand{\miName}{MI\xspace}
\newcommand{\cmiName}{c-MI\xspace}

\newcommand{\wordSimilarity}[0]{w}
\newcommand{\embedingSimilarity}[0]{e}
\newcommand{\wordAggregated}[2]{\wordSimilarity_{#1#2}}
\newcommand{\embeddingAggregated}[2]{\embedingSimilarity_{#1#2}}
\newcommand{\aggregationFunction}[1]{\emph{#1}}
\renewcommand{\sim}[0]{\emph{sim}\xspace}
\newcommand{\dissim}[0]{\emph{dissim}\xspace}
\newcommand{\alphaRange}[0]{\braces{1,0.9,0.8}\xspace}

\newcommand{\aggMax}[0]{\uparrow}
\newcommand{\aggMin}[0]{\downarrow}
\newcommand{\avg}[0]{+}
\newcommand{\product}[0]{\times}

\newcommand{\docCosineAllTest}[0]{\statistic{40.0}}
\newcommand{\docCosineExplicitTest}[0]{\statistic{49.0}}
\newcommand{\docCosineImplicitTest }[0]{\statistic{35.8}}

\newcommand{\docJSImplicitTest }[0]{\statistic{41.2}}

\newcommand{\ngramsAllTest}[0]{\statistic{45.1}}
\newcommand{\ngramsExplicitTest}[0]{\statistic{60.1}}
\newcommand{\ngramsImplicitTest}[0]{\statistic{38.8}}

\newcommand{\simDissimAllTest}[0]{\statistic{31.9}}

\newcommand{\simDissimExplicitTest}[0]{\statistic{52.9}}

\newcommand{\simDissimImplicitTest}[0]{\statistic{31.3}}

\newcommand{\noStemsAllTest}[0]{\statistic{42.1}}

\newcommand{\noStemsExplicitTest}[0]{\statistic{60.8}}

\newcommand{\noStemsImplicitTest}[0]{\statistic{35.3}}

\newcommand{\miAllTest}[0]{\statistic{48.5}}
\newcommand{\miExplicitTest}[0]{\statistic{68.6}}
\newcommand{\miImplicitTest}[0]{\statistic{40.7}}
\newcommand{\cmiAllTest}[0]{\statistic{50.6}}
\newcommand{\cmiExplicitTest}[0]{\statistic{72.5}}

\newcommand{\numAllSpeeches}[0]{\statistic{3685}}
\newcommand{\numAllSOnes}[0]{\statistic{1797}}
\newcommand{\numAllSTwos}[0]{\statistic{1887}}
\newcommand{\numSTwosExplicit}[0]{\statistic{348}}
\newcommand{\numSTwosImplicit}[0]{\statistic{1389}}
\newcommand{\numSTwosImaginary}[0]{\statistic{150}}
\newcommand{\numAllMotions}[0]{\statistic{460}}
\newcommand{\numSOnesWithRebuttal}[0]{\statistic{1102}}
\newcommand{\numMotionsHavingSOneWithRebuttal}[0]{\statistic{329}}
\newcommand{\numSTwoCandidates}[0]{\statistic{1708}}

\newcommand{\averageSentencesPerSpeech}[0]{\statistic{28.2}}
\newcommand{\averageTokensPerSpeech}[0]{\statistic{738.6}}

\newcommand{\numGroupsTrainAll}[0]{\statistic{649}}
\newcommand{\avgCandidatesPerGroupTrainAll}[0]{\statistic{5.2}}
\newcommand{\avgPositivesPerGroupTrainAll}[0]{\percentage{31}}
\newcommand{\numCandidatesTrainAll}[0]{\statistic{1021}}
\newcommand{\numGroupsValidationAll}[0]{\statistic{218}}
\newcommand{\avgCandidatesPerGroupValidationAll}[0]{\statistic{5.2}}
\newcommand{\avgPositivesPerGroupValidationAll}[0]{\percentage{31}}
\newcommand{\numCandidatesValidationAll}[0]{\statistic{340}}
\newcommand{\numGroupsTestAll}[0]{\statistic{235}}
\newcommand{\avgCandidatesPerGroupTestAll}[0]{\statistic{5.2}}
\newcommand{\avgPositivesPerGroupTestAll}[0]{\percentage{31}}
\newcommand{\numCandidatesTestAll}[0]{\statistic{347}}
\newcommand{\numGroupsTrainExplicit}[0]{\statistic{159}}
\newcommand{\avgCandidatesPerGroupTrainExplicit}[0]{\statistic{5.6}}
\newcommand{\avgPositivesPerGroupTrainExplicit}[0]{\percentage{24}}
\newcommand{\numCandidatesTrainExplicit}[0]{\statistic{542}}
\newcommand{\numGroupsValidationExplicit}[0]{\statistic{58}}
\newcommand{\avgCandidatesPerGroupValidationExplicit}[0]{\statistic{6.0}}
\newcommand{\avgPositivesPerGroupValidationExplicit}[0]{\percentage{22}}
\newcommand{\numCandidatesValidationExplicit}[0]{\statistic{208}}
\newcommand{\numGroupsTestExplicit}[0]{\statistic{51}}
\newcommand{\avgCandidatesPerGroupTestExplicit}[0]{\statistic{5.4}}
\newcommand{\avgPositivesPerGroupTestExplicit}[0]{\percentage{25}}
\newcommand{\numCandidatesTestExplicit}[0]{\statistic{188}}
\newcommand{\numGroupsTrainImplicit}[0]{\statistic{556}}
\newcommand{\avgCandidatesPerGroupTrainImplicit}[0]{\statistic{5.0}}
\newcommand{\avgPositivesPerGroupTrainImplicit}[0]{\percentage{30}}
\newcommand{\numCandidatesTrainImplicit}[0]{\statistic{999}}
\newcommand{\numGroupsValidationImplicit}[0]{\statistic{194}}
\newcommand{\avgCandidatesPerGroupValidationImplicit}[0]{\statistic{5.0}}
\newcommand{\avgPositivesPerGroupValidationImplicit}[0]{\percentage{30}}
\newcommand{\numCandidatesValidationImplicit}[0]{\statistic{337}}
\newcommand{\numGroupsTestImplicit}[0]{\statistic{201}}
\newcommand{\avgCandidatesPerGroupTestImplicit}[0]{\statistic{5.1}}
\newcommand{\avgPositivesPerGroupTestImplicit}[0]{\percentage{30}}
\newcommand{\numCandidatesTestImplicit}[0]{\statistic{343}}

\newcommand{\forFinal}[1]{}
\newcommand{\userComment}[2]{}

\usepackage{listings}
\title{Out of the Echo Chamber:\\Detecting Countering Debate Speeches}

\author{
Matan Orbach,
Yonatan Bilu,
Assaf Toledo,
Dan Lahav, 
\\
\textbf{
Michal Jacovi,
Ranit Aharonov 
and
Noam Slonim 
} \\
IBM Research
\\
\{matano,yonatanb,michal.jacovi,noams\}@il.ibm.com\\
\{assaf.toledo,dan.lahav, ranit.aharonov\}@ibm.com
}

\date{}

\begin{document}
\maketitle

\begin{abstract}
An educated and informed consumption of media content has become a challenge in modern times. 
With the shift from traditional news outlets to social media and similar venues, a major concern is that readers 
are becoming encapsulated in "echo chambers" and may fall prey to fake news and disinformation, lacking easy access to dissenting views. We suggest a novel task aiming to alleviate some of these concerns -- that of detecting articles that most effectively counter the arguments -- and not just the stance -- made in a given text. 
We study this problem in the context of debate speeches. 
Given such a speech, we aim to identify, from among a set of speeches on the same topic and with an opposing stance, the ones
that directly counter it. 
We provide a large dataset of \numAllSpeeches such speeches (in English), annotated for this relation, which hopefully would be 
of general interest to the NLP community. We explore several algorithms addressing this task, and while some are successful, all fall short of expert human performance, suggesting room for further research. 
All data collected during this work is freely available for research\datasetURL.
\end{abstract}

\section{Introduction}
Recently, a
publication on Quantum Computing described a quantum computer swiftly performing a task that arguably would require 10,000 years to be solved by a classical computer \cite{arute2019quantum}.
A non-expert reader is likely to consider this claim as a hard-proven fact, especially due to the credibility of the venue in which this publication appeared. Shortly afterwards, a contesting blog written by other experts in that field\footnote{\url{https://www.ibm.com/blogs/research/2019/10/on-quantum-supremacy/}} argued, among other things, that the aforementioned problem can be simulated on a classical computer, using proper optimizations, in $2.5$ days.
Clearly, out of potentially many texts questioning the promise of Quantum Computers (e.g. \citet{kalai2019argument}), 
making readers of the former publication aware of that specific blog post, which directly contests the claims argued in that publication, will provide them with a more informed view on the issue.

Broadly, argumentative texts, such as articles that support a certain viewpoint, often lack arguments contesting that viewpoint.
This may be because those contesting arguments are not known to the author of the text, 
as they might not even have been raised at the time of writing. 
Alternatively, authors may also deliberately ignore certain known arguments, which might undermine their argumentative goal.
Regardless of the reason, this issue places readers at a disadvantage.
Lacking familiarity with opposing views that \emph{specifically challenge a given perspective}, may lead to uninformed decisions or establishing opinions based on partial or biased information. 
Therefore, there is merit to developing a system that can automatically detect such opposing views.

Motivated by this scenario, we propose a novel natural language understanding task: Given an input text and a corpus, retrieve from that corpus a \emph{counter text} which includes arguments contesting the arguments raised in the input text.   
While contemporary systems allow fetching texts on a given topic, and can employ existing tools to discern its stance -- and so identify texts with an opposing view -- 
they lack the nuance to identify the counter text which directly contests the arguments raised in the input text.

The potential use-cases of the proposed system exist in several domains. In politics, it can present counters to partisan texts, thus promoting more informed and balanced views on existing controversies.
In social media, it can alleviate the bias caused by the "echo chamber" phenomenon \cite{Garimella2018echoChambers}, by introducing opposing views.
And in the financial domain, it can potentially help analysts find relevant counter-texts to predictions and claims made in earning calls. 
It may also help authors to better present their stance, by challenging them with counter texts during their writing process. Lastly, it may aid researches to examine relevant citations by annotating which papers, out of potentially many, hold opposing views. Note, however, that this paper focuses on counter text detection - a useful tool for these worthy goals, but not a complete solution.


To pursue the aforementioned task, one needs a corresponding benchmark data, that would serve for training and evaluating the performance of an automatic system. For example, one may start with an opinion article, find a set of opinion articles on the same topic with an opposing stance, and aim to detect those that most effectively counter the arguments raised in the opinion article we started with. 
This path represents a formidable challenge; for example, reliable annotation of long texts is 
notoriously difficult to obtain \cite{lavee-etal-2019-crowd}, to name just one reason out of many. 

To overcome this issue, here we focus on a unique debate setup, in which the goal of one expert debater is to generate a coherent speech that counters the arguments raised in another speech by a fellow debater. 
Specifically, as part of Project Debater\footnote{\url{https://www.research.ibm.com/artificial-intelligence/project-debater/}}, we collected more than 3,600 debate speeches, each around four minutes long, recorded by professional debaters, on a wide variety of controversial topics, posed as debate {\it motions\/} (e.g. \emph{we should ban gambling}). 
With this paper, we make this data available to the community at large.
Each motion has a set of supporting speeches, and another set of opposing speeches, typically recorded in response to one -- and only one -- of the supporting speeches. 
Correspondingly, our task is defined as follows. Given a motion, a supporting speech, and a set of candidate opposing speeches discussing the same motion, identify the opposing speeches recorded in response to the supporting speech. 

We analyze human performance on this challenging task, over a sample of speeches, and further report systematic results of a wide range of contemporary NLP models. Our analysis suggests that expert humans clearly outperform the examined automatic methods, by employing a potentially non-trivial mix of heuristics. 

In summary, our main contributions are as follows:
(1) Introducing a novel NLU task, of identifying the long argumentative text that best refutes a long argumentative text given as input.
(2) Suggesting to simulate the proposed general task in a well-framed debate setup, in which one should identify the response speech(es) that rebuts a given supporting speech.
(3) Sharing a large collection of more than 3,600 recorded debate speeches, that allow to train and evaluate automatic methods in our debate-setup task.
(4) Providing empirical results for a variety of contemporary NLP models in this task.
(5) Establishing the performance of humans in this task, conveying that expert humans currently outperform automatic methods.

\section{Related Work}

Most similar to our work is the task of retrieving the best counter argument to a single given argument \cite{wachsmuth-etal-2018-retrieval}, also within the debate domain.
However, in that setting counterarguments may discuss different motions, or have the same stance towards one motion. 
In our setting, identifying speeches discussing the same motion can be done using 
existing NLP methods, and being of opposing stances may be explored with various sentiment analysis techniques.
Our focus is on identifying the response to a supporting speech 
within a set of \emph{opposing} speeches, \emph{all discussing the same motion}. 
Other than the different setup, our task also 
handles a more complex premise -- speeches which are substantially longer than any single argumentative unit, and include multiple such units.

An alternative to our approach is breaking the problem into three stages: (1) identifying specific arguments made in each debate speech; (2) establishing counterargument relations between such arguments found in different speeches; (3) choosing the best response speech based on these argument-level relations. 
The first sub-problem has been recently explored in \citet{Mirkin-etal:idebate,Lavee2019towardsRebuttal,orbach-etal-2019-dataset}.
The second is related to a major research area within computational argumentation (see recent surveys by \citet{cabrio2018argumentMiningSurvey,lawrence2019argumentationSurvey}).
Such research includes detecting attack relations between arguments \cite{cabrio-villata-2012-combining,rosenthal2015couldn, peldszus2015towards,cocarascu-toni-2017-identifying, wachsmuth-etal-2018-retrieval}, modeling them \cite{sridhar2015joint}, depicting these relations \cite{walker2012corpus, peldszus2015annotated,  musi2017building}, generating counter-arguments \cite{hua-acl2018-rebuttal, hua-acl2019-rebuttal-generation}, and establishing a theoretical framework for engagement \cite{Toulmin:58, Govier1991-GOVAPS, dung95, Damer2009-DAMAFR, Walton09}.

A major drawback of the above approach is that it requires a considerable labeling effort -- the annotation of arguments mentioned within speeches -- which has been shown to be a challenge \cite{lavee-etal-2019-crowd}. 
Another is that the methods in the above studies which focus on establishing relations at the individual argument level may be limited when aiming to evaluate the perspective of long texts. 
Specifically, a response speech may contain multiple arguments that relate to the supporting speech in different ways. 
For instance, the speaker in 
such a speech may choose to concede an argument, while still maintaining an opposite view.
Therefore simply mapping argument level relations may fall short when trying to generalize and assess full speeches.   
Our task complements the above endeavors by facilitating a framework that would allow extending their granularity from the argument level to a full-text level.
Also, our main motivation is different -- detecting whole long counter speeches, and not the exact counter arguments within the counter speech. 
The latter, perhaps more challenging goal, is out of scope for this work.

New neural models have recently driven performance improvements across many NLP tasks \cite{devlin2018bert,Radford-GPT2}, surpassing the level of non-expert humans in a diverse set of benchmark tasks \cite{wang-etal-2018-glue, Decathlon}. To facilitate the progress of further research \citet{wang2019superglue} introduced a benchmark aiming to pose a new series of rigorous tests of language understanding which are challenging for cutting-edge NLP technologies. 
Our work is consistent with the motivation behind these benchmarks, as it suggests a challenging new NLU task, accompanied by a corresponding dataset and benchmarks.

The rise of deliberate disinformation, such as fake news, highlights the erosion in the credibility of consumed content  \cite{fakenews:18},
and situations where one is exposed only to opinions that agree with their own, as captured by the notion of echo chambers, are becoming more prevalent \cite{Garimella2018echoChambers, duseja-jhamtani-2019-sociolinguistic}.
The task proposed in this work seems timely in this context.

\section{Data}
\label{sec:data}

We now detail the process of collecting the speeches, the structure of the dataset, and how it is used for our task.

\begin{figure}[t]
\centering
\includegraphics[trim={6.55cm 11.4cm 13.85cm 1.5cm},clip,width=75mm]{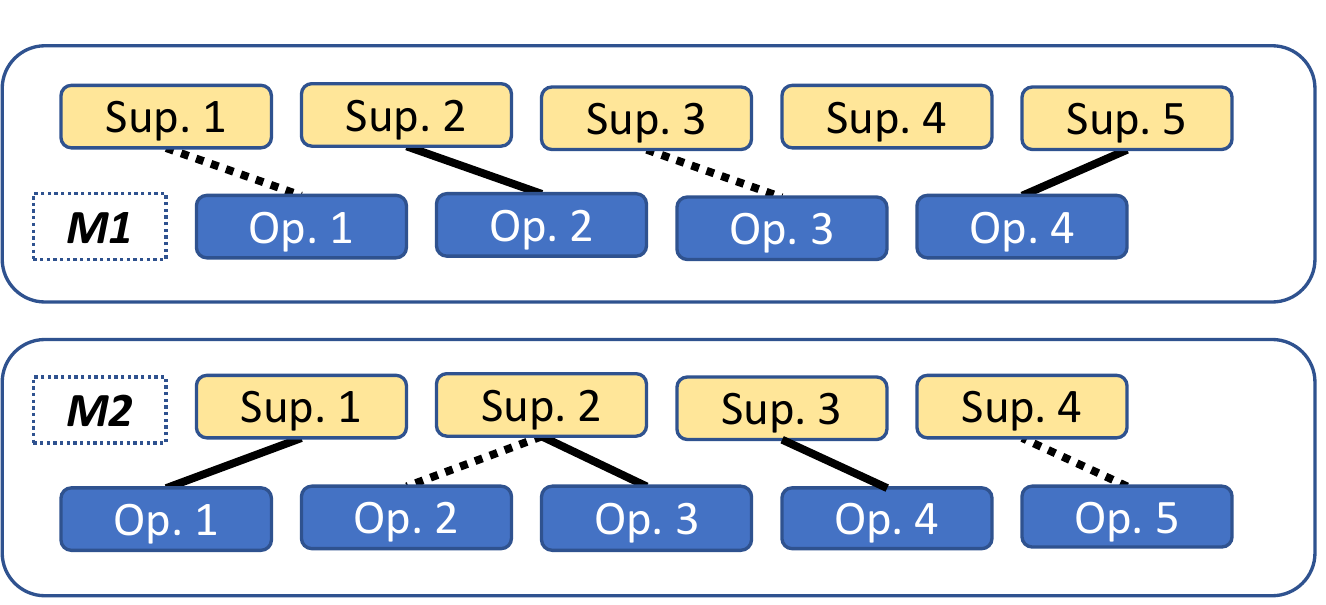}
\caption{
The speeches data structure for two motions (\emph{\textbf{M1}} and \emph{\textbf{M2}}): 
Each motion has several supporting (\textbf{Sup.}) and opposing (\textbf{Op.}) speeches.
Opposing speeches which constitute an \textbf{explicit}/\textbf{implicit} counter speech to a supporting speech are connected to it with a \textbf{solid}/\textbf{dashed} line. 
In the data, each supporting speech has zero or more counters, and each opposing speech is the counter of at most one supporting speech.\label{fig:speeches-structure}}
\end{figure}

\paragraph{Dataset structure}
Each speech in the dataset discusses a single motion and is either a \emph{supporting speech} -- in which a single speaker is arguing in favor of the discussed motion, or an \emph{opposing speech} -- in which the speaker is arguing against the motion, 
typically in response to a supporting speech for that motion. 
As described below, debaters recording 
an opposing speech typically
listen to a given recorded supporting speech, and then design and record their own speech in response to it. 
This \emph{counter speech} is
either \emph{explicit} -- including a rebuttal part in which the speaker directly addresses arguments raised in the rebutted speech, or \emph{implicit} -- 
including no such dedicated rebuttal section, but tacitly relating to the issues raised in the supporting speech they respond to. The data contains multiple counter speeches to each supporting speech, among which some, none or all may be explicit or implicit.
\figureRef{fig:speeches-structure} depicts the structure of this dataset. 
Examples of one explicit and one implicit counter speeches are included in the \supp.

\paragraph{Recording speeches}
The supporting speeches were produced by a team of professional debaters, using a procedure similar to the one described in \citet{Mirkin-etal:2018}:
The debaters were each given a list of motions, accompanied by relevant background materials (taken from an online resource such as Wikipedia). 
They were allowed ten minutes of preparation time 
to review a motion's background material, after which they recorded a speech arguing in favor of that motion,
which was around four minutes long. 
Through this process, \numAllSOnes supporting speeches were recorded, discussing \numAllMotions 
motions.

To record an opposing speech, the debaters were first given ten minutes to review the background material for the motion, as in the recording of a supporting speech.
Then, they listened to a supporting speech (recorded by a fellow debater) and recorded a counter 
speech of similar length.
Due to different debate styles popular in different parts of the world, some debaters recorded explicit counter speeches while others recorded implicit ones. 
To expedite the pace of the recording process, towards its end, few opposing speeches were recorded without requiring the debater to respond to a specific supporting speech. 
Instead, the debaters were instructed to think of supporting arguments themselves, and respond to these arguments. 
In total, \numAllSTwos opposing speeches were recorded: \numSTwosExplicit are explicit counters,  \numSTwosImplicit are implicit, and the other  \numSTwosImaginary are not the counter speech of any supporting speech.
The full guidelines used by the debaters during the recordings are included in the \supp.

The recorded audios were automatically transcribed into text using Watson's off-the-shelf Automatic Speech to Text (STT)\footnote{\url{https://www.ibm.com/cloud/watson-speech-to-text}}. 
Human transcribers listened to the recorded speeches, and manually corrected any errors found in the transcript texts produced by the STT system.
On average, each speech transcript contains
\averageSentencesPerSpeech sentences, and averages \averageTokensPerSpeech tokens in length.

For the purpose of this work, the manually-corrected transcripts are used. 
The full data of \numAllSpeeches speeches, including the recorded audios, the STT system outputs and the manually-corrected transcripts are available on our website\datasetURL. 
For comparison, the previous release of Project Debater's speeches dataset \cite{Lavee2019towardsRebuttal} included a smaller subset of $400$ speeches. 
Further details on the format of the full data and the recordings process are available in \citet{Mirkin-etal:2018}. 

\paragraph{Usage}
As noted above, our task input is comprised from a supporting speech and several candidate opposing speeches all discussing the same motion.
Some candidates are counters of the supporting speech, and others are typically counters of \emph{other} supporting speeches for the same motion.
The goal is to identify those counter speeches made in response to the supporting speech.
Opposing speeches produced by the speaker of the supporting speech were excluded from the candidates set, as in the real world it is unexpected for one to simultaneously support both sides of a discussion.

\section{Human Performance}
\label{sec:annotation}
Recently, with deep learning techniques achieving human performance on several NLU tasks, and even surpassing it, there is growing interest in raising the bar \cite{wang2019superglue}. That is, 
to facilitate advancing NLU beyond the current state-of-the-art, there is a need for novel tasks which are solvable by humans, yet challenging for automatic methods.
To assess our proposed task in this context, we performed an annotation experiment, as described below. 

\paragraph{Setup} Each question presented one supporting speech and between 3 to 5 candidate opposing speeches, all discussing the same motion.
Annotators were instructed to read the speeches, and select one opposing speech which they thought was a counter speech of the supporting speech.
When they could not identify such a counter, they were asked to guess and mention that they had done so.

\annotationQuestionsAll questions were randomly sampled and given to \statistic{3} English-proficient expert annotators, who have successfully worked with our team in other past annotation experiments.
Following their choice of a counter speech, they were asked to explain their choice in free form language.

Following this step, one of the authors read the explanations provided by the experts and formed a set of reason categories. 
Then, another \annotationQuestionsAll questions were sampled and given to \statistic{3} crowd annotators, using the  
\textit{\figureEightName}\footnote{\url{www.figure-eight.com}}
crowdsourcing platform. 
The crowd annotators were from a dedicated group which regularly participates in annotations done by our team.
After choosing a counter speech, they were instructed to choose the reason (or multiple reasons) for their choice from the set of reason categories. The crowd payment was set to $2.5\$$ per question. To encourage thorough work, a post-processing bonus was given for each correct answer, doubling that pay. 

The full guidelines given to the expert and crowd annotators are provided in the \supp.

\paragraph{Results} 
Performance was evaluated by calculating the accuracy of each annotator, and averaging over annotators.
These results are presented in \tableRef{tab:annotation_results}. 
Overall, the experts obtained an average accuracy of \allInternalPrecisionPercentage (\textbf{\expertsRowName} row), considerably better than randomly guessing the answer which yielded an accuracy of \allInternalRandomPrecisionPercentage.
The accuracy of the crowd annotators (\textbf{\crowdRowName}) was lower, yet distinctly better than random. This suggests that the task is difficult, and may require a level of dedication or expertise beyond what is common for crowd-annotators. Fortunately, the dataset is constructed in such a way that human annotation is not required to label it - it is clear by design which opposing speech counters which supporting speech.

\begin{table}[t]
\begin{center}
\begin{tabular}{lcccccc}
\toprule
& \multicolumn{2}{c}{\bf \emph \allSpeechesCriterion} 
& \multicolumn{2}{c}{\bf \emph \explicitSpeechesCriterion} 
& \multicolumn{2}{c}{\bf \emph \implicitSpeechesCriterion}
\\
\cmidrule(rl){2-3}
\cmidrule(rl){4-5}
\cmidrule(rl){6-7}
&
\accuracyTitle & \randomTitle &
\accuracyTitle & \randomTitle &
\accuracyTitle & \randomTitle
\\
\midrule
\bf \expertsRowName & 
\allInternalPrecision &
\allInternalRandomPrecision &
\explicitInternalPrecision  &
\explicitInternalRandomPrecision  &
\implicitInternalPrecision &
\implicitInternalRandomPrecision 
\\
\bf \crowdRowName & 
\allCrowdPrecision &
\allCrowdRandomPrecision &
\explicitCrowdPrecision  &
\explicitCrowdRandomPrecision &
\implicitCrowdPrecision &
\implicitCrowdRandomPrecision 
\\
\bottomrule
\end{tabular}
\end{center}
\caption{Annotation results showing, for each \emph{\textbf{annotation setting}},  
the average accuracy (\accuracyTitle) obtained by the experts (\textbf{\expertsRowName}) and crowd annotators (\textbf{\crowdRowName}), along with the accuracy of randomly guessing the answer (\randomTitle).}
\label{tab:annotation_results} 
\end{table}

To establish whether identifying explicit counters is easier than identifying implicit ones, the average annotator accuracy was separately computed for these two types.
Noteworthy, the accuracy of the experts drops from a near perfect score of \explicitInternalPrecisionPercentage on questions with an explicit true counter, to \implicitInternalPrecisionPercentage on questions with an implicit one.
Some of the drop may be explained by the smaller chance of guessing the correct answer at random over this set, but not all\footnote{Suppose that when answering, annotators answer correctly a fraction $f$ of the time, and guess $1-f$ of the time, with probability of success equal to the random baseline.
Then in the explicit case $f=0.87$ and in the implicit $f=0.67$.}. 
This suggests that, as may be expected, identifying implicit counter speeches is more challenging than identifying an explicit counter. 
Still, the performance of both types of annotators, over both types of speeches, was better than random.

\paragraph{Reasons analysis} 
The explanations provided by the experts revealed several best-practices for this task, which we categorized as follows:
The true counter speech \textbf{quote}s a phrase from the supporting speech;
\textbf{mention}s a specific case or argument from the supporting speech;
is more \textbf{comprehensive} and addresses more issues raised in the supporting speech than the other candidates; addresses those issues in the same \textbf{order} as they appear in the supporting speech; discusses \textbf{similar} issues; deals with the \textbf{main issue} raised in the supporting speech.
Another reason was \textbf{elimination} -- discarding the other candidates since they responded to issues or arguments which were \emph{not} raised in the supporting speech.
The last two categories were \textbf{guess} and \textbf{other} (which required writing a reason in free form language).

\begin{figure}[t]
\centering
\includegraphics[trim={2.05cm 9.8cm 2cm 8.8cm},clip,width=75mm]{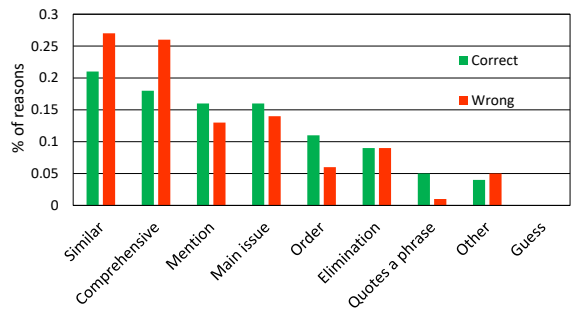}
\caption{The distribution of reasons for the correct and wrong answers of crowd annotators (who overall had accuracy $\geq$ $\percentage{60}$).
\label{fig:reasons-histogram}
}
\end{figure}

Focusing on crowd annotators who did the task relatively well (accuracy $\geq$ $\percentage{60}$),
\figureRef{fig:reasons-histogram} presents the distribution of the reasons they gave for their answers, separated between cases when they were correct and when they were wrong.
Overall, the reasons distribution suggests that correctly solving this task requires balancing between the various heuristics.
While some of these reasons, such as \textbf{similarity}, correspond to existing algorithmic ideas, others (e.g. \textbf{order} or \textbf{main issue})
could inspire future research.

\section{Counter Speech Identification}
\label{sec:experiments}

Having established that experts perform well on this task, the question remains whether present NLP methods can match that performance.

\subsection{Setup}
\label{subsec:experimental_setup}

\paragraph{Data}A supporting speech was included in the experiments if (a) there was an opposing speech addressing it; and (b) there was at least one additional opposing speech discussing its motion  which was 
produced either in response to \emph{another} supporting speech, or without responding to any specific supporting speech.
Supporting speeches not meeting these criteria were excluded from the analysis.
With these criteria, 
the data used in the experiments comprised \numSOnesWithRebuttal supporting speeches and \numSTwoCandidates opposing speeches, pertaining to \numMotionsHavingSOneWithRebuttal motions.

\paragraph{Split}The motions were randomly split into \trainNameLower (\percentage{60}), \validationNameLower (\percentage{20}) and \testNameLower (\percentage{20}) sets, and their speeches were partitioned accordingly.

\paragraph{Settings}To separately evaluate the ability to detect explicit and implicit counters, the experiments were performed in three settings. 
The first utilized the entire data -- given a supporting speech, all of the opposing speeches discussing its motion were considered as candidate counters.  
In the second setting, the true counter speeches were limited to explicit counters.
Supporting speeches without any explicit counter were excluded.
Similarly, in the last setting, the true counter speeches were limited to implicit counters, and supporting speeches without such counters were excluded.
For example, a supporting speech with one explicit counter, one implicit counter and whose motion is associated with two other opposing speeches (which are not its counters), is considered with all four opposing speech candidates in the first setting and three such candidates in the second and third settings - the two non-counters and the one counter of the type relevant to the setting.
\tableRef{tab:split_stats} details the statistics of each data split and experimental setting.

\begin{table*}[t]
\begin{center}
\begin{tabular}{lcccccccccccc}

\toprule
&
\multicolumn{4}{c}{\bf \trainName} &
\multicolumn{4}{c}{\bf \validationName} &
\multicolumn{4}{c}{\bf \testName} 
\\
\cmidrule(rl){2-5}
\cmidrule(rl){6-9}
\cmidrule(rl){10-13}
\emph{\textbf{Setting}}&
\supportingSpeechTitle & 
\numCandsTitle & 
\averageCandsTitle &
\percentPositiveTitle &
\supportingSpeechTitle & 
\numCandsTitle & 
\averageCandsTitle &
\percentPositiveTitle &
\supportingSpeechTitle & 
\numCandsTitle & 
\averageCandsTitle &
\percentPositiveTitle 
\\ 
\midrule
\bf \emph \allSpeechesCriterion
&
\numGroupsTrainAll & 
\numCandidatesTrainAll &
\avgCandidatesPerGroupTrainAll &
\avgPositivesPerGroupTrainAll &
\numGroupsValidationAll & 
\numCandidatesValidationAll & 
\avgCandidatesPerGroupValidationAll &
\avgPositivesPerGroupValidationAll &
\numGroupsTestAll & 
\numCandidatesTestAll & 
\avgCandidatesPerGroupTestAll &
\avgPositivesPerGroupTestAll 
\\
\bf \emph \explicitSpeechesCriterion
&
\numGroupsTrainExplicit & 
\numCandidatesTrainExplicit & 
\avgCandidatesPerGroupTrainExplicit &
\avgPositivesPerGroupTrainExplicit &
\numGroupsValidationExplicit & 
\numCandidatesValidationExplicit &
\avgCandidatesPerGroupValidationExplicit &
\avgPositivesPerGroupValidationExplicit &
\numGroupsTestExplicit & 
\numCandidatesTestExplicit &
\avgCandidatesPerGroupTestExplicit &
\avgPositivesPerGroupTestExplicit 
\\
\bf \emph \implicitSpeechesCriterion
&
\numGroupsTrainImplicit & 
\numCandidatesTrainImplicit & 
\avgCandidatesPerGroupTrainImplicit &
\avgPositivesPerGroupTrainImplicit &
\numGroupsValidationImplicit & 
\numCandidatesValidationImplicit & 
\avgCandidatesPerGroupValidationImplicit &
\avgPositivesPerGroupValidationImplicit &
\numGroupsTestImplicit & 
\numCandidatesTestImplicit &
\avgCandidatesPerGroupTestImplicit &
\avgPositivesPerGroupTestImplicit 
\\
\bottomrule
\end{tabular}
\end{center}
\caption{Data split statistics for each \emph{\textbf{experimental setting}} (see \sectionRef{subsec:experimental_setup}): The number of supporting (\supportingSpeechTitle) and opposing (\numCandsTitle) speeches, the average number of candidate counter speeches for each supporting speech (\averageCandsTitle), and the percentage of those candidates which are counter speeches (\percentPositiveTitle).}
\label{tab:split_stats} 
\end{table*}

\paragraph{Evaluation}
The methods described next score each of the candidate counters. We report the average accuracy of the top predictions (\precisionTitle) and the average mean reciprocal rank (\mrrTitle), defined as $1/r$ where $r$ is the highest rank of a true counter.

\subsection{Methods}

\paragraph{Document similarity}
Our first method represented speeches as bag-of-terms vectors, where 
terms are stemmed unigrams appearing
in at least \percentage{1} of the speech-pairs in the training set, and the term counts are normalized by the total count of terms in the speech. 
Given two vectors, their similarity was computed using the Cosine similarity (\emph{\docSimCosine}) or the inverse Jensen-Shannon divergence (\emph{\docSimJS}). 

\paragraph{Similarity and Dissimilarity}
\citet{wachsmuth-etal-2018-retrieval} presented a method for retrieving the best counter argument to a given argument, based on capturing the similarity and dissimilarity between an argument and its counter. 
At its core, their method is based on two similarity measures between pairs of texts:
(i) A \emph{word-based} similarity, which is defined by the inverse 
Manhattan distance 
between the normalized term frequency vectors of the texts (where terms were as mentioned above);
(ii) An \emph{embeddings-based} similarity which used pretrained ConceptNet Numberbatch word embeddings \cite{speer2017conceptnet} to represent the words of the texts, 
averaged those embeddings to obtain a vector representing each text,
and calculated the inverse Word Mover's distance \cite{Kusner:2015:WED:3045118.3045221} between these vectors.

Previously, these measures were used to predict the relations between a pair of argumentative units.
Since our speeches may contain multiple arguments, and their location within the text is unknown, 
we defined this method at the speech level by considering
every supporting speech sentence 
and every candidate counter speech sentence.
For each measure, the similarities of one supporting speech sentence to all candidate counter speech sentences were aggregated by applying a function \aggregationFunction{f}, yielding a sentence-to-speech similarity.
These sentence-to-speech similarities were aggregated using another function \aggregationFunction{g}, yielding a speech-to-speech similarity. We denote these speech-to-speech measures by $\wordAggregated{f}{g}$ for word-based similarities and $\embeddingAggregated{f}{g}$ for embedding-based similarities.
As aggregation functions, the maximum ($\aggMax$), minimum ($\aggMin$), average ($\avg$) and product ($\product$) were considered.
For example, $\wordAggregated{\aggMax}{\avg}$ denotes taking the maximal word-based similarity of each supporting speech sentence to all candidate counter speech sentences, and averaging those values.

Lastly, following \citet{wachsmuth-etal-2018-retrieval} once more, the similarity (\emph{\simDissimName}) between a supporting speech and a candidate counter is defined as 
\[
\alpha \cdot \sim - (1-\alpha) \cdot \dissim
\] where \sim and \dissim are of the form $\wordAggregated{f}{g}+\embeddingAggregated{f}{g}$, both \aggregationFunction{f} and \aggregationFunction{g} are aggregation functions, 
$\sim\neq\dissim$ and $\alpha$ is a weighting factor. 
In this scoring model \sim aims to capture topic similarity, whereas subtracting \dissim seeks to capture the dissimilarity between arguments from opposing stances. 
Admittedly, this method is more appropriate for some of the settings explored in \citet{wachsmuth-etal-2018-retrieval}, in which the candidate counter arguments to a given argument may be discussing other topics, and their stance towards the discussed topic is unknown. We include their method here for completeness, and to allow a comparison to their work.

The hyper-parameters, namely, the aggregation functions and the value of $\alpha$ 
(from the range $\alphaRange$ used by \citet{wachsmuth-etal-2018-retrieval}) 
were tuned on the validation set.
An additional variant (\emph{\simDissimEmbeddingsName}) based solely on the embeddings-based similarity was also considered, since it carries the advantage of not requiring any vocabulary to be derived from the training set.
This allowed tuning the hyper-parameters on a larger set comprised from both the training and validation sets.

\paragraph{BERT} 
\citet{devlin2018bert} presented the BERT
framework which was pre-trained on the masked language model and next sentence prediction tasks. 
Assuming that an argument and its counter are coherent as consecutive sentences, and that the first sentences of the candidate speech reference the last sentences of the supporting speech, those parts were scored using the pre-trained next-sentence prediction model with (\textit{\tunedBert}) and without (\textit{\vanilaBert}) fine-tuning.
The considered sentences from each speech were limited to at most $100$ words, since 
the pre-trained model is limited to $512$ word pieces (assuming about two word pieces per word). 
Specifically, from the first speech we took the greatest number of sentences from the end of the speech such that their total length was less than $100$ words, and similarly for the second speech for its starting sentences. For fine-tuning, we used the supporting speeches with each of their true counter speeches as positive sentence pairs, and added an equal number of negative pairs where the supporting speech appears with a randomly sampled opposing speech that is not its counter.

\paragraph{ngram-based}
The methods described so far assign a score to a supporting speech and a candidate counter without considering the other candidates.
Using that content can aid in detecting key phrases or arguments which best characterize the connection between the supporting speech and its counter -- 
these are the ones which are shared between those speeches and are not mentioned in any of the other candidates.
Having many such phrases or arguments may be an indication that a candidate is a true counter speech. 
Indeed, the \textbf{quote} and \textbf{mention} reason categories account for 
more than \percentage{20} of the reasons selected by the crowd annotators when answering correctly (see \tableRef{fig:reasons-histogram}).

To capture this intuition, ngrams containing between $2$ to $4$ tokens were extracted from each speech.
Those containing stopwords, and those fully contained within longer ngrams, were removed.
The set of ngrams which appear in both the supporting speech and the candidate -- but not in any of the other candidates -- was calculated, and the total length of the ngrams it contains was used as the score of the candidate (\emph{\ngramScoreName}). 

\paragraph{Mutual Information}
The speeches were represented as bag-of-terms binary vectors, where the terms are stemmed unigrams (excluding stopwords). 
Each candidate counter was scored using the mutual information between its vector and the vector of the supporting speech (\emph{\miName}).

In addition, the mutual information between those vectors, conditioned by the presence of terms in the other candidate counters (\emph{\cmiName}), was calculated as follows.
Let $v_{s}$ be a vector representing a supporting speech and $\braces{v_{c}}_{c=1}^{n}$ be a set of $n$ vectors representing its candidate counters.
Let $c$ be such a candidate counter, and $o_{c}$ represent the concatenation of the vectors of the \emph{other} candidates excluding $c$.
Let $\vk{c}$ denote the vector of values from $v_{c}$ at the indices where the entries of $o_{c}$ are $k$ (for $k=1$ or $0$) , and let $\vk{s}$ be defined similarly. 
Then, the conditional mutual information of the candidate $c$ is given by
\[
    \sum_{k=0}^{1} \pk I(\vk{s};\vk{c})
\]
where $\pk$ is the percentage of entries of $o_c$ with the value $k$, and $I(\cdot,\cdot)$ is mutual information.
Intuitively, this measure aims to quantify the information shared between a 
supporting speech and a candidate, 
after observing the content of all other candidates, and thus is similar in spirit to the ngram-based method mentioned above.

\subsection{Results}
\tableRef{tab:test_results} presents the results obtained by the different methods in our three experimental settings. 
These results show that there is a large performance gap  between the implicit and explicit settings -- in favor of the latter -- for all methods (except \emph{\vanilaBert}), 
suggesting it is an easier setting.
This is consistent with the results of our annotation experiment.

While the best performing methods (\emph{\docSimJS} and \emph{\cmiName}) surpass the performance of individual crowd annotators (see \tableRef{tab:annotation_results}), which testifies to the difficulty of the annotation task, the human experts clearly do better, suggesting there is still much room for improvement.

\begin{table}[t]
\begin{center}
\tabcolsep=0.118cm
\begin{tabular}{lcccccc}
\toprule
& \multicolumn{2}{c}{\bf \emph \allSpeechesCriterion} 
& \multicolumn{2}{c}{\bf \emph \explicitSpeechesCriterion} 
& \multicolumn{2}{c}{\bf \emph \implicitSpeechesCriterion}
\\
\cmidrule(rl){2-3}
\cmidrule(rl){4-5}
\cmidrule(rl){6-7}
\textbf{Method} & 
\precisionTitle & \mrrTitle &
\precisionTitle & \mrrTitle &
\precisionTitle & \mrrTitle 
\\
\midrule
\bf \docSimJS & 
$\mathbf{51.1}$ &
$\mathbf{0.69}$ &
$\mathbf{80.4}$ &
$\mathbf{0.88}$ &
\docJSImplicitTest &
\statistic{0.62} 
\\
\bf \cmiName &  \cmiAllTest & 
$\mathbf{0.69}$ &
\cmiExplicitTest & 
\statistic{0.84} &
$\mathbf{42.7}$ &
$\mathbf{0.63}$
\\
\bf \miName &  \miAllTest & 
\statistic{0.68} &
\miExplicitTest & 
\statistic{0.81} &
\miImplicitTest & 
\statistic{0.62} 
\\
\bf \ngramScoreName & 
\ngramsAllTest & 
\statistic{0.65} &
\ngramsExplicitTest & 
\statistic{0.73} &
\ngramsImplicitTest &
\statistic{0.61} 
\\
\bf \simDissimEmbeddingsName  & 
\noStemsAllTest & 
\statistic{0.63} &
\noStemsExplicitTest & 
\statistic{0.76} &
\noStemsImplicitTest &
\statistic{0.58}
\\
\bf \docSimCosine & 
\docCosineAllTest & 
\statistic{0.62} &
\docCosineExplicitTest & 
\statistic{0.70} &
\docCosineImplicitTest &
\statistic{0.58} 
\\
\bf \vanilaBert &
\statistic{36.4} & 
\statistic{0.57} &
\statistic{21.6} & 
\statistic{0.44} &
\statistic{33.7} &
\statistic{0.57} 
\\
\bf \simDissimName & 
\simDissimAllTest & 
\statistic{0.57} &
\simDissimExplicitTest & 
\statistic{0.70} &
\simDissimImplicitTest &
\statistic{0.56}
\\
\bf \tunedBert &
\statistic{32.2} & 
\statistic{0.56} &
\statistic{49.0} & 
\statistic{0.70} &
\statistic{35.2} &
\statistic{0.58} 
\\ 
\midrule
\bf Rand & 
\percentage{31} & \statistic{-} &
\percentage{25} & \statistic{-} &
\percentage{30} & \statistic{-} 
\\ 
\bottomrule

\end{tabular}
\end{center}
\caption{Experimental results on the test set
for each \textbf{method} and \emph{\textbf{experimental setting}}:
The average accuracy of the top prediction (\precisionTitle) and the average mean reciprocal rank (\mrrTitle) of the true counter with the highest score. The methods are ordered by their \mrrTitle\xspace score in the \textit{All} setting. The last row (\textbf{Rand}) shows the accuracy of the random baseline.}
\label{tab:test_results} 
\end{table}

\paragraph{Error analysis} 
We have manually analyzed the top 3 implicit and explicit speeches for which the differences in mutual information between the predicted counter speech and the true counter speech were the greatest.
Analysis revealed that such counter speeches are characterized by argumentative material that is thematically similar to the material of the input speech. 
Depending on the use case, such results are not necessarily errors, since if the goal is to find relevant opposing content it is beneficial to present
such speeches, even if they were not authored in response to the input speech. 
However, in some instances a thematically similar argument may be an irrelevant counter as arguments can share a theme without being opposing.
For example, an input text may discuss an argument pertaining to the \emph{rights} of a disenfranchised group, while the counter may revolve around \emph{pragmatic outcomes} to the same disenfranchised group. While these arguments are likely to share the theme of disenfranchisement they are not necessarily opposing.   
\section{Further Research Potential}
\label{sec:usecases}

The data presented here was collected to facilitate the development of Project Debater, and we chose the novel counter speech detection task to showcase this data and make it available to the community. However, the unique properties of our data -- recorded speech which is more organized and carefully construed than everyday speech -- make it interesting to revisit well-known NLP and NLU tasks. 
Several examples are listed below.

\paragraph{Author attribution:} All speeches 
in the dataset are annotated for the debater who recorded them. It could be particularly interesting to study author attribution on our dataset as it contains persuasive language, relevant to opinion writing and social media. 
Additionally, we provide voice recordings and transcripts for all speeches, 
enabling to study multi-modal methods for this task.
\paragraph{Topic identification:} This is a well established research area which can be examined here in various aspects, including clustering speeches by topic, matching speeches to topics or extracting the topic of a speech without prior knowledge. 

Whereas previous work often requires annotating the topics of texts and deducing a consensual label, in our data the topic of a speech is given by design.

\paragraph{Sentence ordering or local coherence:} The sentence ordering task  \cite{barzilay-lapata-2005-modeling} is concerned with organizing text in a coherent way and is especially relevant for natural language generation. 
Our dataset allows to study this using spoken natural language of a persuasive nature, that often relies on a careful development of an argumentative intent. The data also provides a unique opportunity to study the interplay between a coherent arrangement of language and the associated prosodic cues.

\paragraph{Other tasks} The large scale of the dataset, over $200$ hours of spoken content and their manually-corrected transcripts, enables its use in other speech-processing tasks that require such data. 
Some examples include speech-to-text, text-to-speech, and direct learning from speech of word \cite{chung2018speech2vec} or sentence \cite{haque2019audio} embeddings.
Such tasks often use large scale datasets of \emph{read} content (e.g. \citet{librispeeech2015}), and our data allows their exploration in the context of spoken \emph{spontenous} speech.

In addition, with further annotations of the dataset, it lends itself to other potential tasks.
One example is the extraction of the main points of a speech or article. 
This can facilitate various downstream tasks, such as single document summarization in the context of spoken language.
Another example is the annotation of named entities within the transcript texts, facilitating direct identification of those entities in the audio, similarly to the work of \citet{ghannay2018nerOnSpeech}.
\section{Conclusions}

We presented a novel NLU task of identifying a counter speech, which best counters an input speech, within a set of candidate counter speeches. 

As previous studies have shown, and consistent with our own findings, obtaining data for such a task is difficult, especially considering that labeling at scale of full speeches is an arduous effort. 
To facilitate research of this problem, we recast the proposed  general task in a defined debate setup and construct a corresponding benchmark data. We collected, and release as part of this work, more than 3,600 debate speeches annotated for the proposed task. 

We presented baselines for the task, 
considering a variety of contemporary NLP models. The experiments suggest that the best results are achieved using Jensen–Shannon similarity, for speeches that contain explicit responses
(accuracy of \percentage{80}) and using conditional mutual-information on speeches that respond to the input speech in an implicit way
(accuracy of \percentage{43}).

We established the performance of humans on this task, showing that expert humans currently outperform automatic methods by a significant margin --- attaining an accuracy of $92\%$ on speeches with an explicit true counter, and $76\%$ on speeches with an implicit one.
Noteworthy is that some of the automatic methods outperform the results achieved by the crowd, suggesting that the task is difficult, and may require a level of expertise beyond layman-level. 

The reported gap between the performance of expert humans and the results achieved by NLP models demonstrate room for further research. Future research may focus on the motivation we described, but may also utilize the large speeches corpus we release as part of this work to a variety of additional different endeavors. 

\section*{Acknowledgments}
We wish to thank the many debaters and transcribers that took part in the effort of creating this dataset, and the anonymous reviewers for their insightful
comments, suggestions, and feedback.

\bibliographystyle{resources/acl_natbib}
\bibliography{main}

\clearpage
\newpage
\appendix

\section{Introduction}
This appendix contains the guidelines used in all the data generation and annotation tasks described in the paper: 1) speech authorship guidelines, 2) identifying the response speech from a list of candidates, 3) identifying the response speech speech from a list of candidates and providing a reason.

Following the guidelines are two examples of full response speeches - an explicit counter speech and an implicit counter speech (see \sectionRef{sec:data}).

\section{Speech Authoring Guidelines}
For supporting speeches:
\begin{itemize}
\item Read the motion text and background.
\item Prepare for 10 minutes while avoiding the use of external sources.
\item Record a ~4 min opening speech in a normal speaking pace.
\end{itemize}

For opposing speeches:
\begin{itemize}
\item Read the motion text and background.
\item Prepare for 10 minutes while avoiding the use of external sources.
\item Listen to the supporting speech.
\item Immediately record a ~4 min opening speech in a normal speaking pace.
\item When recording your speech, please make sure to relate to the arguments raised in the government's opening speech; i.e., engage with them like you would have done in British Parliamentary debate style, or in any other kind of academic debate format. 
\end{itemize}

\section{Identify The Opposing Speech Guidelines}
In this task you are given a motion and speech arguing in favor of that motion.
It is then followed by 3-5 opposing speeches. One of those speeches was recorded in response to the first supporting speech.
Please select the opposing speech which you think was recorded in response to the supporting speech. In addition, please write in your own language the reason for your choice.

Note that you MUST select exactly one opposing speech. If you aren’t sure, take a guess, and specify you had done so when detailing the reason for your choice.
Some additional examples of valid reasons are “Both X and Y seemed reasonable choices, and X seemed more appropriate”, “The supporting speech is talking about Z, as does the opposing speech”, etc.
No specific format is required for detailing the reason, but please do your best to be clear.

\section{Identify The Opposing Speech (With Reasons) Guidelines}
\subsection*{Overview}
In this task you are given a controversial topic and a supporting speech arguing in favor of that topic.
The supporting speech is followed by 3-5 opposing speeches. 
One of those opposing speeches was recorded in response to the supporting speech.

\begin{enumerate}
\item Select the opposing speech that was recorded in response to the supporting speech. 
\item Select the reasons for your choice from a predefined list of reasons. You can select more than one reason.
\item Explain your choice, in your own words, in case the reason for your choice does not appear in the list.
\end{enumerate}

Note that you MUST select exactly one opposing speech. 
If you aren’t sure, take a guess, and specify you had done so when selecting the reason for your choice from the predefined list.
When explaining your choice in your own words, no specific format is required  -- but please do your best to be clear.

\subsection*{Important Note}
This task does not contain test questions, but your answers will be reviewed after the job is complete. We trust you to perform the task thoroughly, while carefully following the guidelines.

\section{Example Speeches}
\subsection*{Explicit counter speech: Opposing subsidies for higher education}
"Before we begin there is something that, at least to me, was remained unclear in the mechanism, and that is the question of what exactly is going to get subsidized and what isn't.
Do liberal arts studies or humanities studies are they going to get the same full funding like computer science or engineering?
We think that this is important because no matter what the answer is going to be, this raises some serious questions and difficulties but anyway, we're going to put that aside for now in the hope that government will make this clear in the next speech.
So, side government is asking to convince us in the following things: a, education, no matter what age, is a basic right.
B, if there is a basic right, then this automatically means that the government is also responsible to fully fund this.
C, subsidizing, like a full subsidy of higher education, is going to be a smart investment that pays off in the long run, both economically and socially.
We disagree with literally every one of those stages.
Let's explain why.
Firstly, on education in every age being a basic right.
So government basically start by saying: look, we can all agree that primary education is a basic right and therefore, we must agree that that higher education is also a basic right.
Now that is a logical leap.
There are plenty of protections and special rights that we provide children but not adults.
Children are protected, for instance, from criminal liability.
And according to government's logic, if that is true, then this should also apply to adults.
This is of course absurd.
Specifically, the line that we cross between primary education to higher education isn't at all random.
Primary education is a crucial condition to succeed in life, no matter what field you're going to to find yourself in.
And that's what makes it a basic right.
It is also a tool of the state to create a shared basis of knowledge to all of the citizens, sort of a way to shape the shared narrative and the collective identity of the nation.
Higher education, on the other hand, isn't a crucial condition in plenty in like a lot of fields and and frankly, in the previous years, it is becoming less and less critical for success.
In addition, there is also no element of like a a shared foundation here because everybody studies different things entirely, so no, this is not a basic right.
Secondly, even if we were to agree that this is a basic right, this doesn't automatically mean that the government need to completely fully fund it.
Food is also a basic right, right?
And still the state helps you very partially and does not provide food for everyone free.
We need to say this very clearly.
The state already participates today in the funding of higher education in public institutes but in a partial way.
We think that demanding that it will provide for all of it is simply a misguided way of perceiving what the state's role is.
Why isn't it enough to fund scholarships for less well-off students and continue collecting money from students that have no problem to fund themselves, for instance?
And lastly, we get to the question of whether this is a smart investment.
Now, as I have already hinted, higher education might have been critical for success in the market ten years ago or fifteen years ago, but the market is rapidly changing today and more and more of the most desired job places, for instance, in google or facebook, don't even demand a an academic title.
We think that before we run off to spend billions of dollars on higher education free for everybody, then it's worth at least heavily considering these institutional changes, and that is something that side government isn't even considering.
For all these reasons, please oppose."

\subsection*{Implicit counter speech: Opposing disbanding ASEAN}
"We should not disband ASEAN.
So, ASEAN is the association of southeast asian states.
As the last speaker pointed out to you, it's made up of a group of states in southeast asia who are working together towards common goals of development.
Three reasons why we should not disband it.
First is about anti-colonialism.
Recognize that for developing countries like the ones in ASEAN like malaysia, like indonesia, they have a few alternatives for who they can turn to as trade partners.
You have major international trading countries like the states, like china, like EU countries, which historically have treated these countries in a very colonialist way.
Most of the countries in ASEAN except for thailand were once colonized by european countries or the united states, and if you look back before that, they had a semi colonial relationship with china in many instances, such the relationship between vietnam and china.
So we see that there's a history of abuse and mistreatment between these larger countries around the world the more powerful countries, and the ASEAN countries.
We think that by working together, the ASEAN countries can ensure that they are a large enough economic bloc to prevent these major international powers who have historically come in and pushed them around, from dominating the region, in other words, ASEAN makes all of these countries that together are strong and able to resist imperialist aggression or trade policy, and would all individually not be that powerful.
It allows them to work together towards a common goal of independence and it reassures the independence of every member state from international oppression and dependence on one country for trade.
Our second argument, is about why we think that fundamentally ASEAN increases development and that's the highest good in this round.
So first, why is development the most important good?
If you think about the quality of life in ASEAN countries, obviously it varies.
People in malaysia for instance have like a middle income quality of life, people in vietnam are much poorer, but we think that overall everyone in all of these countries could still benefit from more development.
We think that there is a moral imperative for states to seek out development for their citizens.
Why is this so?
So when we say a moral imperative we mean that states always have an obligation to seek this out.
We think that because, any person would always choose to live in the most developed country possible so that they have the highest quality of life, those with the ability to do so, those who reap the benefits of developed life, because they're elites, should try to provide it to everyone else.
To sort of do unto others as you would have them do unto you type of thinking.
We see that, development is more likely with ASEAN.
One, because countries have more access to trade partners and trade goods, so it's more like that they're able to specialize and develop industries that can then take advantage of other markets within ASEAN, and two, because of the access to economic development expertise.
Recognize that many countries in ASEAN, are at different levels of development.
Malaysia is pretty far along, some other countries are not as far along.
So we tell you that people in ASEAN countries can study in other countries and learn about development and industry, and how other countries have been successful in the past, and use this in order to help their own home country.
So at the end of the day, we help the people who are worse off in the world, some of them, some of these very poor people who live in ASEAN countries because we better access development so we shouldn't disband ASEAN.
Our last argument is about peace in the region.
Recognize that there are many potential sources of conflict within the southeast asia region.
Some countries are more closely aligned with china so they see an advantage in china becoming more hegemonic, some countries are more aligned with the united states.
Some countries are communists, some countries are capitalist.
There's been conflict in the past over east timor, and there are other ethnic tensions boiling beneath the surface in many southeast asian countries.
But one of the surest ways to prevent international conflict, is to tie everyone's interests together through trade.
If everyone stands to get richer through peace and poorer through conflict, then it's much less likely that a war will break out in the region.
So for that reason we think ASEAN is a tremendous tool for peace in southeast asia in the future.
So because it's an anti colonial institution, because it promotes development, and because it will lead to peace in the region, we should not disband ASEAN thank you."

\end{document}